\documentclass[bdcc,article,accept,pdftex,moreauthors]{Definitions/mdpi} 

\firstpage{1} 
\pubvolume{1}
\issuenum{1}
\articlenumber{0}
\pubyear{2025}
\copyrightyear{2025}
\externaleditor{Firstname Lastname} 
\datereceived{ } 
\daterevised{ } 
\dateaccepted{ } 
\datepublished{ } 
\hreflink{https://doi.org/} 

\usepackage[most]{tcolorbox}

\Title{TACO: Adversarial Camouflage Optimization on Trucks to Fool Object Detectors}

\TitleCitation{TACO: Adversarial Camouflage Optimization on Trucks to Fool Object Detectors}

\Author{Adonisz 
 Dimitriu $^{1,}$*$^{,\dagger}$, Tamás Vilmos Michaletzky 
 $^{1,\dagger}$ and Viktor Remeli $^{2}$}

\AuthorNames{Adonisz Dimitriu, Tamás Vilmos Michaletzky and Viktor Remeli}

\AuthorCitation{Dimitriu, A.; Mihaletzky, T.; Remeli, V.}

\address{%
$^{1}$ \quad Technology Transfer Institute, 1113 Budapest, Hungary 
; michaletzky.tamas@techtra.hu 
\\
$^{2}$ \quad Széchenyi István University  
, 9026 Győr, Hungary 
; remeli.viktor@techtra.hu} 

\corres{Correspondence
: dimitriu.adonisz@techtra.hu} 

\firstnote{These authors contributed equally to this work.}

\abstract{Adversarial attacks threaten the reliability of machine learning models in critical applications like autonomous vehicles and defense systems. As object detectors become more robust with models like YOLOv8, developing effective adversarial methodologies is increasingly challenging. We present Truck Adversarial Camouflage Optimization (TACO), a novel framework that generates adversarial camouflage patterns on 3D vehicle models to deceive state-of-the-art object detectors. Adopting Unreal Engine 5, TACO integrates differentiable rendering with a Photorealistic Rendering Network to optimize adversarial textures targeted at YOLOv8. To ensure the generated textures are both effective in deceiving detectors and visually plausible, we introduce the Convolutional Smooth Loss function, a generalized smooth loss function. Experimental evaluations demonstrate that TACO significantly degrades YOLOv8's detection performance, achieving an AP@0.5 of 0.0099 on unseen test data. Furthermore, these adversarial patterns exhibit strong transferability to other object detection models such as Faster R-CNN and earlier YOLO~versions.}

\keyword{adversarial attacks; object detection; camouflage optimization; AI security; YOLOv8; Unreal Engine 5} 

\begin{document}

\begin{tcolorbox}[colback=yellow!10!white, colframe=gray!50!black, boxrule=0.5pt, width=\textwidth]
\centering
\small\textbf{This is the accepted version of the article published in Big Data and Cognitive Computing (MDPI). Please cite the final version at \url{https://doi.org/10.3390/bdcc9030072}}
\end{tcolorbox}

\section{Introduction}
In recent years, object detection has made significant advances, with~the YOLO family of models leading the way in real-time applications. These models have become essential in areas like autonomous driving, surveillance, and~robotics, offering high accuracy and efficiency in many applications~\cite{yolov1-8}. However, as~these technologies become more integrated into critical systems, their vulnerabilities have also come to light. Adversarial attacks, which involve subtle, often imperceptible perturbations, have been shown to cause machine learning models to make incorrect predictions, exposing significant security risks~\cite{goodfellow}. While these attacks initially focused on classifiers, the~scope has since expanded, with~adversarial methods now being applied to domains such as object detection, image segmentation~\cite{Adv_semantic_seg}, reinforcement learning~\cite{Adv_RL}, and even large language models~\cite{Adv_LLM}. 

Adversarial attacks can be broadly categorized into digital and physical attacks. Digital attacks involve manipulating pixel values in an image to fool a model. These attacks are effective when the input to the model is a digital image, but~they fail when applied to real-world objects. For~instance, a~pixel change that deceives an object detector digitally may become ineffective when printed or viewed under different lighting and perspectives, as~changes in illumination or camera angles can alter the pixel values of a printed~pattern.

This limitation has driven the development of physical adversarial attacks, where the patterns are physically applied to objects. These attacks must remain effective across varying lighting conditions, angles, and~environmental factors. Recent research has demonstrated that such adversarial attacks are feasible. For~example~\cite{shadow_attack}, has shown that even natural phenomena, like certain shadows, can serve as effective adversarial attacks, revealing how deceivable object detection models can be. Similarly,~\cite{background_attack} introduced a novel approach that focuses on background adversarial perturbations in both digital and physical domains, showing that Deep Neural Networks (DNNs) can be deceived by perturbations applied to the background rather than the objects~themselves.

To effectively craft physical adversarial patterns, a~differentiable image generation pipeline is essential. Such a pipeline enables the optimization of adversarial patterns by allowing gradients of a loss function—typically tied to the object detection model's confidence score—to propagate through the entire rendering process. 
This capability is critical because it allows the gradients to flow seamlessly from the output of the object detector back to the 2D texture applied to the 3D model. By~doing so, the~optimization process can directly adjust the texture to minimize the detection confidence score or alter the detector's behavior. 
Ultimately, a~successful pipeline ensures that the generated patterns are specifically tuned to exploit the vulnerabilities of the object detector, effectively camouflaging the object against detection systems.

In this study, we introduce Truck Adversarial Camouflage Optimization (TACO), a~novel framework designed to render a specific truck model undetectable to state-of-the-art object detection models by generating adversarial camouflage patterns. Leveraging Unreal Engine 5 (UE5) for photorealistic and differentiable rendering, TACO optimizes textures applied to a 3D truck model to deceive detectors, specifically targeting YOLOv8~\cite{yolov8_ultralytics}. Our fully differentiable pipeline integrates advanced rendering techniques with neural networks to optimize adversarial patterns that prevent the detection of the truck.
The key contributions of TACO are as follows:

\begin{itemize}
    \item We are the first to utilize UE5 for generating adversarial patterns within a fully differentiable rendering pipeline. This advancement builds upon prior methods that employed Unreal Engine 4 (UE4), offering improved graphical fidelity and rendering capabilities. 
    {Using UE5 we reduce the domain gap between the simulated environment and real-world deployment, ensuring adversarial patterns remain effective in the real world.
    }
    \item We introduce an additional neural rendering component, a~gray textured truck image, to accurately capture and reproduce lighting and shadow conditions.
    \item We are the first to design adversarial patterns specifically for YOLOv8 in the context of vehicle detection, moving beyond previous work that focused on older models like YOLOv3~\cite{yolov3}. 
    \item {
    We introduce Intersection over Prediction-based (IoP-based) filtering as part of the class loss formulation, enhancing the stealthiness of adversarial optimization by considering bounding boxes that significantly overlap with the target object. This method reduces false detections and improves the overall effectiveness of adversarial patterns.
    }
    \item We propose the Convolutional Smooth Loss function, a~novel smooth loss function for ensuring that the adversarial textures are not only effective but also visually plausible.
\end{itemize} 

The rest of the paper is organized as follows. Section~\ref{section2} explores related works on adversarial attacks using visual patterns. Section~\ref{section3} formulates the problem statement and the TACO framework. Section~\ref{section4} shows the implementation details, followed by a presentation of our results in Section~\ref{section5}. Section~\ref{section6} concludes the~paper.  

\section{Related~Works}\label{section2}
Adversarial attacks on object detection systems have garnered significant attention in recent years. While early research primarily focused on digital adversarial examples, the~shift towards physical-world attacks has introduced new challenges and methodologies. This section reviews the evolution of physical adversarial attacks, particularly those targeting vehicles, and~highlights how our work advances the state of the~art.

One of the initial approaches to physical adversarial attacks involved the use of adversarial patches. For~instance, it was shown that attaching patches to specific regions in an image could deceive object detectors into misclassifying or failing to detect objects~\cite{DPatch}. Building on this concept, it was also demonstrated that holding a printed adversarial patch in front of a person could successfully evade person detection systems~\cite{Patch_infront}. While these studies provided valuable insights, they primarily focused on human subjects and simple~scenarios.

Shifting the focus to vehicle-based applications, which are particularly relevant for autonomous driving and the military sector, researchers explored new methods to deceive object detectors. An~innovative approach was to attach a screen to a car that displays adversarial patterns dynamically adjusted based on the camera's viewpoint~\cite{patch_car_screen}. Although~this method proved effective, it relies on electronic displays and knowledge of the detector camera location, which may not be practical or covert in real-world~scenarios.

To overcome the limitations of screen-based methods, a~black-box method was proposed to approximate both rendering and gradient estimation for generating adversarial patterns~\cite{CAMOU}. Notably, they observed that increasing the resolution of the camouflage does not necessarily enhance the fooling rate of object detectors, suggesting a trade-off between pattern complexity and~effectiveness.

Further advancements were made by exploring white-box attacks that leverage knowledge of the target model's parameters. Two primary strategies emerged for constructing differentiable pipelines for pattern optimization. The~first strategy involved projecting a pattern onto the surface of the object using camera parameters. For~example, a~differentiable transformation network approach (DTA) projected repeated patterns onto vehicles~\cite{DTA}. However, this approach suffered from projection errors, especially on non-planar surfaces, leading to inaccuracies in the application of adversarial~patterns.

Addressing these limitations, triplanar mapping was introduced (ACTIVE~\cite{ACTIVE}), which projects the pattern from three different planes to reduce distortion on complex geometries. 
{
However, it still struggles to maintain texture proportions on complex geometries, such as bent or highly curved surfaces, because~the mapping does not account for the underlying surface’s spatial distortion. 
}

An alternative and more accurate approach involves the use of neural mesh renderers, also known as differentiable renderers~\cite{diff_renderer}. 
By mapping textures directly onto each triangle face of the 3D mesh using UV coordinates, differentiable renderers ensure that every pixel in the final rendered view aligns with the underlying geometry, allowing reliable gradient backpropagation into the texture space and minimizing artifacts. The Dual Attention Suppression (DAS) approach~\cite{DAS} utilized this technique, aiming to suppress the attention maps of the target detection model while generating visually natural adversarial textures relying on partial coverage using~patches. 

\newpage
However, partial coverage is less effective compared to full-body textures; the superior performance of the Full-coverage Camouflage Attack (FCA) was demonstrated by Wang~et~al.~\cite{FCA}, achieving increased robustness and performance in deceiving object detectors. Building upon this, Zhou~et~al. proposed RAUCA where they further enhanced the effectiveness of adversarial patterns by simulating various weather conditions, such as different times of day, rain, and~fog~\cite{RAUCA}. They incorporated environmental information from background images, which they passed through their Environment Feature Extractor network. Their results indicated that patterns optimized under diverse scenarios exhibit greater resilience in real-world~applications.

In parallel, Duan~et~al. proposed a method to generate adversarial patterns for a 3D truck object, specifically targeting the Faster R-CNN object detector~\cite{CAC}. Their approach combined 3D rendering with dense proposal attacks to train adversarial camouflage across varying viewpoints and lighting conditions, further advancing the effectiveness of 3D adversarial attacks on vehicle-based~models.

Recent advancements have explored alternative methodologies for generating adversarial patterns. For~instance, Li~et~al. introduced diffusion models to create adversarial textures~\cite{flexible}, moving away from traditional optimization-based approaches. Similarly, Lyu~et~al. proposed a novel framework that leverages diffusion models to generate customizable and natural-looking adversarial camouflage patterns for vehicle detectors~\cite{CNCA}. By~allowing users to specify text prompts, their method produces diverse and more natural textures while maintaining competitive attack~performance.

However, despite these advancements, several challenges remain in crafting effective physical adversarial attacks against state-of-the-art object detectors. Many prior works, such as FCA~\cite{FCA}, DAS~\cite{DAS}, DTA~\cite{DTA}, and~ACTIVE~\cite{ACTIVE}, have relied on UE4. While effective, UE4 does not achieve the level of photorealism offered by its successor, UE5. The advanced features of UE5, such as Lumen for real-time global illumination, Nanite for handling detailed geometry, and~an enhanced physically-based rendering pipeline, enable the generation of textures that more accurately replicate real-world lighting, shadows, and~material properties. This increased photorealism ensures that adversarial patterns remain effective when transferred from a simulated environment to physical deployment.
Additionally, these studies often targeted older versions of object detection models like YOLOv3 or Faster R-CNN, which may not reflect the robustness of current state-of-the-art detectors~\cite{yolov1-8}.

RAUCA achieved progress by extracting features from background images to enhance the realism of adversarial patterns. However, their method may not accurately capture the complex interplay of lighting and shadows directly on the~vehicle.

Our work addresses these gaps by utilizing UE5 within a fully differentiable rendering pipeline, enabling the generation of more photorealistic images and detailed adversarial patterns. Furthermore, we target YOLOv8, a~state-of-the-art object detection model known for its robustness and improved detection capabilities. By~introducing an additional neural rendering component—a gray textured truck image—we accurately capture environmental lighting and shadows cast on the vehicle, further improving the accuracy of our neural~renderer.

\begin{figure}[H]
    
    \includegraphics[width=\textwidth]{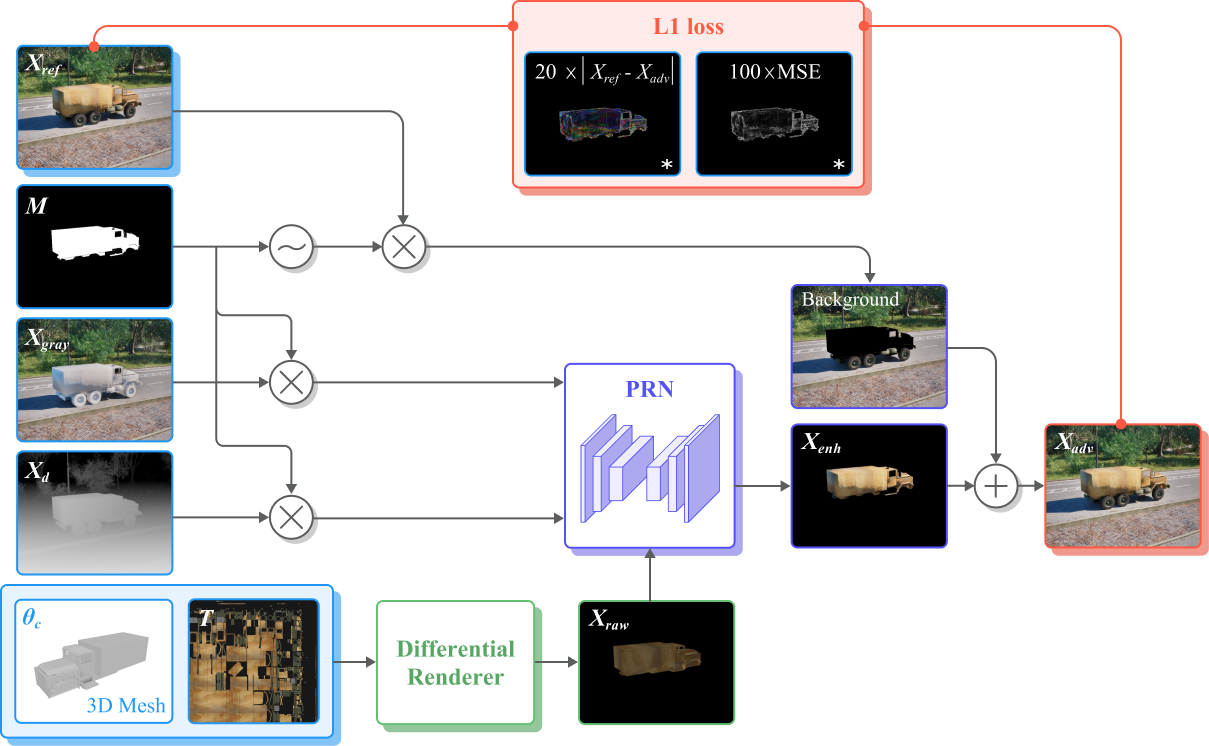}
    \caption{Data 
 flow and training process of our neural renderer, composed of two elements: a differential renderer and the Photorealistic Rendering Network (PRN). The~error images (absolute difference and mean squared error, magnified) are marked with an asterisk (*) to indicate that they are included for illustration purposes~only.}
\label{fig:neural_renderer}
\end{figure}
\unskip

\section{Materials and~Methods}\label{section3}
\unskip
\subsection{Problem~Statement}

Let $ X = \{ X_{1}, X_{2}, \dots, X_{n} \} $ be the dataset generated in UE5, where each sample $ X_{i} $ includes the following elements:     

\begin{itemize}
    \item \textbf{Reference Image 
 ($X_{ref}$)}: A photorealistic image of the truck in the scene, rendered with UE5’s advanced lighting and shading techniques. The~texture for this image is randomly sampled from a High-Resolution Texture Dataset defined in Section~\ref{sec:dataset}.
    \item  \textbf{Gray Textured Truck Image ($X_{gray}$)}: A version of the truck rendered in a neutral gray texture (RGB:127,127,127).
    \item \textbf{Depth Map ($X_d$)}: A depth map that provides the distance from the camera to each pixel on the truck surface.
    \item \textbf{Binary Mask ($M$)}: A mask identifying the pixels corresponding to the truck in each image. It is generated using a custom material in UE5, which ensures black color during rendering for accurate segmentation.
    \item \textbf{Camera Parameters ($\theta_c$)}: Parameters that define the camera’s pose and orientation in the scene, used as input for differential rendering.
\end{itemize}
In 
addition to these components, we define the 3D truck model with its associated mesh and texture map $T$. The~adversarial texture $ T_{adv} $ is used in place of $ T $ to deceive the detection model. The~neural renderer $ R $ takes as inputs the 3D mesh, adversarial texture $ T_{adv} $, camera parameters $ \theta_c $, depth map $ X_d $, and the gray textured truck image $ X_{gray} $ to render the photorealistic adversarial truck image $ X_{enh} $:
\begin{equation}
X_{enh} = R(Mesh, T_{adv}, \theta_c, X_d, X_{gray}).
\end{equation}

Next, the~rendered truck image $ X_{enh}$ is combined with the background $X_{ref} $ using the binary mask $ M $:
\begin{equation}
\label{eq:X_adv}
X_{adv} = X_{enh} \cdot M + X_{ref} \cdot (1 - M),
\end{equation}
where $ X_{enh} \cdot M $ \textls[-21]{places the truck into the scene and $ X_{ref} \cdot (1 - M) $ fills the rest of the~background}. 

It is important to note that our primary goal is to make the truck undetectable. Object detection models, including YOLOv8, output bounding boxes with coordinates $B_{pred} = (b_x, b_y, b_w, b_h)$ and a confidence score $b_{cls}$ for the object class. 
For many real-world applications, such as surveillance and autonomous systems, the~presence of an object (vehicle or a person) is more critical than its exact location. Therefore, our main objective is to minimize the detection confidence score $b_{cls}$ for all classes. 
The optimization problem can be formulated as:
\begin{equation}
T^*_{adv} = \underset{T_{adv}}{\arg \min} \ \mathcal{L}(F(X_{enh}; \theta_F))
\end{equation}
\textls[-5]{where $F$ is the detector with parameters $\theta_F$ and $\mathcal{L}(·)$ is the loss function designed to reduce the class confidence score $ b_{cls}$. This loss function, along with a smoothness regularization term for the adversarial texture $T_{adv}$, is detailed in later sections (see \mbox{Sections~\ref{sec:attack_loss} and ~\ref{sec:smooth_loss}}).}

\subsection{Truck Adversarial Camouflage~Optimization}

\subsubsection{Neural Renderer: Overview of the Rendering~Process}

A central component of the TACO framework is the neural renderer, a~system designed to generate photorealistic images of a truck adorned with adversarial camouflage patterns. This component ensures that the textures are rendered on the 3D object in a fully differentiable and photorealistic~manner.

The rendering process consists of two main stages. First, adversarial textures are applied to the truck's 3D mesh using a differentiable renderer, referred to as $R_{diff}$. This renderer, given the truck mesh, adversarial texture $T_{adv}$, and~camera parameters $\theta_c$, outputs a raw rendered image, $X_{raw}$:
\begin{equation} 
X_{raw} = R_{diff}(\text{Mesh}, T_{adv}, \theta_c).
\end{equation}

In the second stage, the~raw image $X_{raw}$ is passed to the Photorealistic Rendering Network (PRN), an~enhancement module that improves the quality of the rendered image by incorporating additional information such as the gray textured truck image $X_{gray}$ and the depth map $X_d$. The~enhanced image $X_{enh}$ is computed as:
\begin{equation} 
X_{enh} = R_{PRN}(X_{raw}, X_{gray}, X_d),
\end{equation}
where $R_{PRN}$ represents the~PRN.

\textls[-21]{Finally, the~enhanced truck image $X_{enh}$ is blended with the background using Equation~\eqref{eq:X_adv}.} The entire data flow and training process of the neural renderer, including both the differentiable renderer and the PRN, is illustrated in Figure~\ref{fig:neural_renderer}.

\subsubsection{Photorealistic Rendering Network (PRN)}
\begin{figure}[H]
    
    \includegraphics[width=0.95\textwidth]{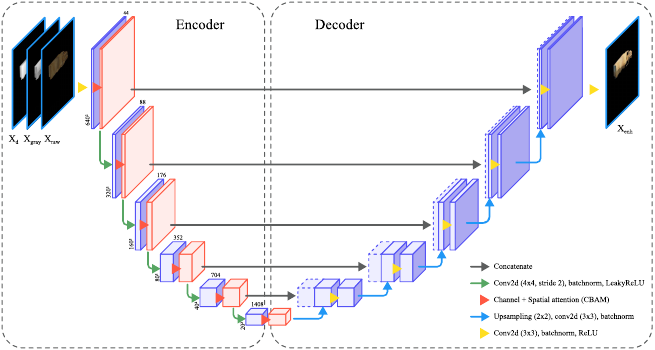}
    \caption{Architecture of the Photorealistic Rendering Network (PRN) based on a U-Net. The~contracting path extracts features, while the expansive path reconstructs the photorealistic image. CBAM modules are used in the contracting path for attention~\cite{cbam}.} 
    \label{fig:prn_architecture}
\end{figure}

{
The PRN is central to transforming raw rendered images into photorealistic outputs. Built on a U-Net architecture~\cite{UNet}, it features a contracting path for feature extraction and an expansive path for image reconstruction. Each downsampling step in the contracting path incorporates a Convolutional Block Attention Module (CBAM) \cite{cbam}, leveraging spatial and channel attention mechanisms to enhance feature representation. The~architecture of the PRN is illustrated in Figure~\ref{fig:prn_architecture}, showing how the contracting and expansive paths interact to produce high-fidelity outputs.
}

{
During training, the~PRN learns the photorealistic rendering characteristics of Unreal Engine 5 (UE5) by using a dataset of reference images $X_{ref}$ with diverse textures. The~training is guided by an L1 rendering loss:
}
\begin{equation}
\mathcal{L}_{render} = || X_{adv} - X_{ref} ||_1,
\end{equation}
{
which ensures that the PRN's output maintains fidelity to $X_{raw}$ while incorporating the photorealistic characteristics of the UE5 reference dataset. For~more details on the dataset and training details refer to Sections~\ref{sec:dataset} and~\ref{sec:prn_training}.
}

\begin{figure}[H]
    
    \includegraphics[width=0.98\textwidth]{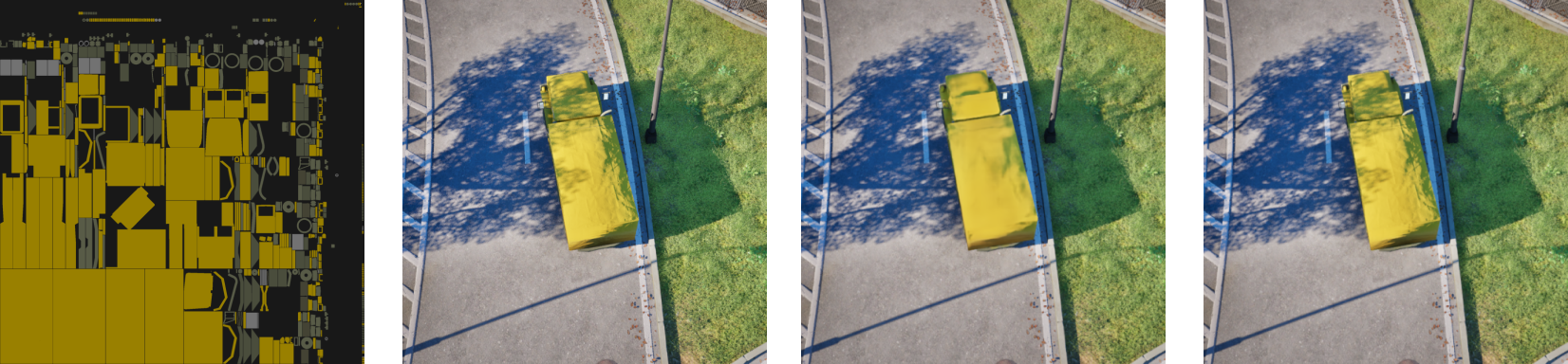}
    \caption{Comparison of shadow rendering quality with and without the gray textured truck input. The figure shows, from~left to right: (1) 
 applied texture, (2) ground truth rendering from UE5, (3) output of the neural renderer excluding $X_{gray}$, and~(4) output of the neural renderer including the $X_{gray}$. Shadows cast on the truck are poorly rendered without $X_{gray}$, but~are accurately captured when it is~included.}
    \label{fig:shadow_comparison}
\end{figure}

{
A key innovation of our approach is incorporating a gray textured truck image~$X_{gray}$ as an additional input to the PRN. This gray textured image captures vital environmental details, particularly lighting and shadows on the vehicle, that would otherwise be difficult to reproduce. As~shown in Table~\ref{tab:shadow_comparison}, models trained with $X_{gray}$ achieve lower L1 loss and higher Strutural Similarity Index Measure (SSIM) \cite{ssim} than those trained without it. Likewise, Figure~\ref{fig:shadow_comparison} illustrates that omitting $X_{gray}$ leads to inaccurate and incomplete shadow rendering, whereas including it yields flawless results to the eye.
}
\begin{table}[H]
    \centering
    \caption{Comparison of PRN performance with and without gray textured truck~input.}
    \begin{tabularx}{\textwidth}{LCC}
        \toprule
        \textbf{Configuration} & \textbf{L1 Loss} & \textbf{SSIM} \\
        \midrule
        Without $X_{gray}$ & 0.031 & 0.9862 \\
        With $X_{gray}$ & \textbf{0.025 
} & \textbf{0.9901} \\
        \bottomrule
    \end{tabularx}
    \label{tab:shadow_comparison}
\end{table}
\vspace{-9pt}

\unskip

\subsubsection{Attack~Loss}
\label{sec:attack_loss}
\begin{figure}[H]
    
    \includegraphics[width = \textwidth]{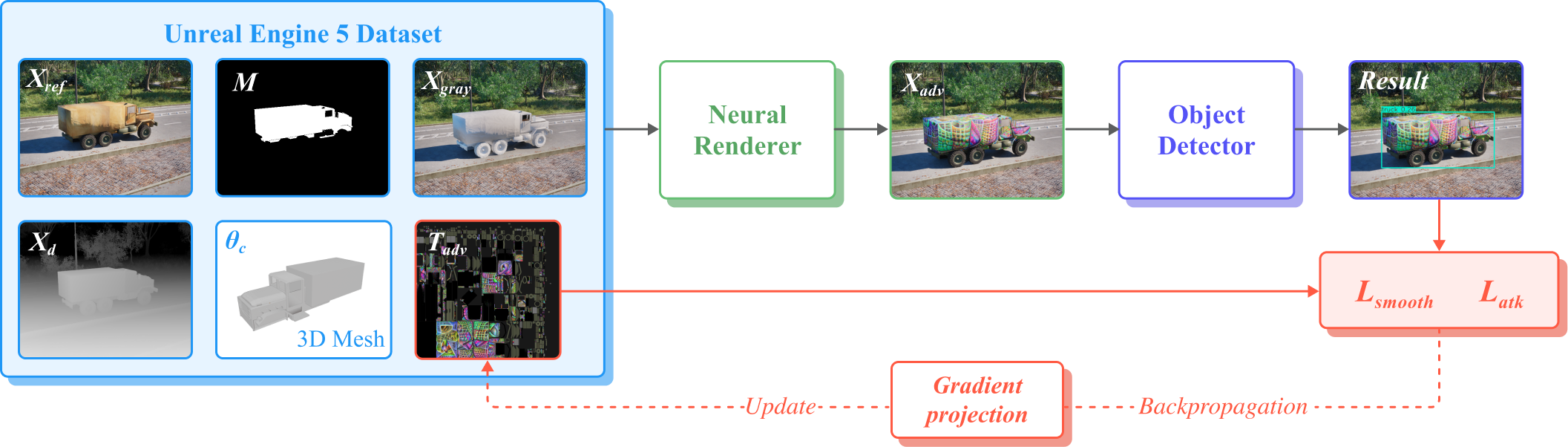}
    \caption{Texture optimization~framework.}
    \label{fig:texture_optimization}
\end{figure}

Following the pre-training of the neural renderer, we now focus on the optimization of the adversarial texture (Figure~\ref{fig:texture_optimization}) 
. In~our setup, we target the Ultralytics YOLOv8 model, which features an updated detection head. Unlike previous YOLO versions, this updated head is anchor-free and removes the objectness score as it was found to be redundant. Instead, the~model directly outputs class confidence scores $b_{cls}$ for~each object in the scene. All YOLO versions in our experiments (e.g., YOLOv3u, YOLOv5Xu, see Section~\ref{sec:evaluation_models}) utilize this updated detection head from Ultralytics, which significantly increases their performance~\cite{new_head}.
Our primary goal is to minimize $b_{cls}$ values specifically for bounding boxes that overlap with the truck. Additionally, we also reduce the Intersection over Union (IoU) between any predicted bounding boxes and the ground truth bounding box of the vehicle. We found that this helps to further reduce $b_{cls}$.

Let $B_{gt}$ denote the ground truth bounding box of the truck and~$B_{pred}^i$ denote the $i$-th predicted bounding box. To~identify predicted bounding boxes that overlap significantly with the truck, we define the Intersection over Prediction (IoP): 
\begin{equation} 
    \text{IoP}(B_{pred}^i,B_{gt}) = \frac{\text{Area}( B_{pred}^i\cap B_{gt})}{\text{Area}(B_{pred})} 
\end{equation}
We select only those predicted bounding boxes where the IoP exceeds a predefined threshold $\tau_{IoP}$ 
. This filtering ensures that our attack focuses on reducing the detection confidence of bounding boxes representing the truck while avoiding patterns that might be misclassified as another class.
The class loss is then defined as
\begin{equation} 
\mathcal{L}_{cls} = - \sum_{c = 1}^C \sum_{i \in \Omega_{IoP}} \log(1 - b_{cls}^{i,c}) 
\end{equation}
where:
\begin{itemize}
    \item $\Omega_{IoP} = \{ i \mid \text{IoP}(B_{pred}^i, B_{gt}) > \tau_{IoP} \}$ denotes the set of indices for bounding boxes with an IoP greater than the threshold $\tau_{IoP}$;
    \item $b_{cls}^{i,c}$ is the confidence score for class $c$ in bounding box $i$;
    \item $C$ is the number of classes (80 in the case of YOLOv8).
\end{itemize}

{
We use IoP-based filtering instead of the traditional IoU-based techniques used in previous works~\cite{ACTIVE, RAUCA}. Our experiments revealed that IoU filtering \(\bigl(\mathrm{IoU} = \tfrac{\mathrm{Area}(B_{pred} \cap B_{gt})}{\mathrm{Area}(B_{pred} \cup B_{gt})}\bigr)\) often excluded bounding boxes covering smaller central regions of the truck when their IoU with the ground truth box $B_{gt}$ fell below the threshold. This exclusion led to adversarial patterns that caused the detector to misclassify parts of the truck as unrelated objects, such as apples or suitcases. Although~the truck itself became undetected, these false identifications of surface patterns compromised the attack’s overall stealth and effectiveness.
}

\begin{figure}[H]
    
    \includegraphics[width=\textwidth]{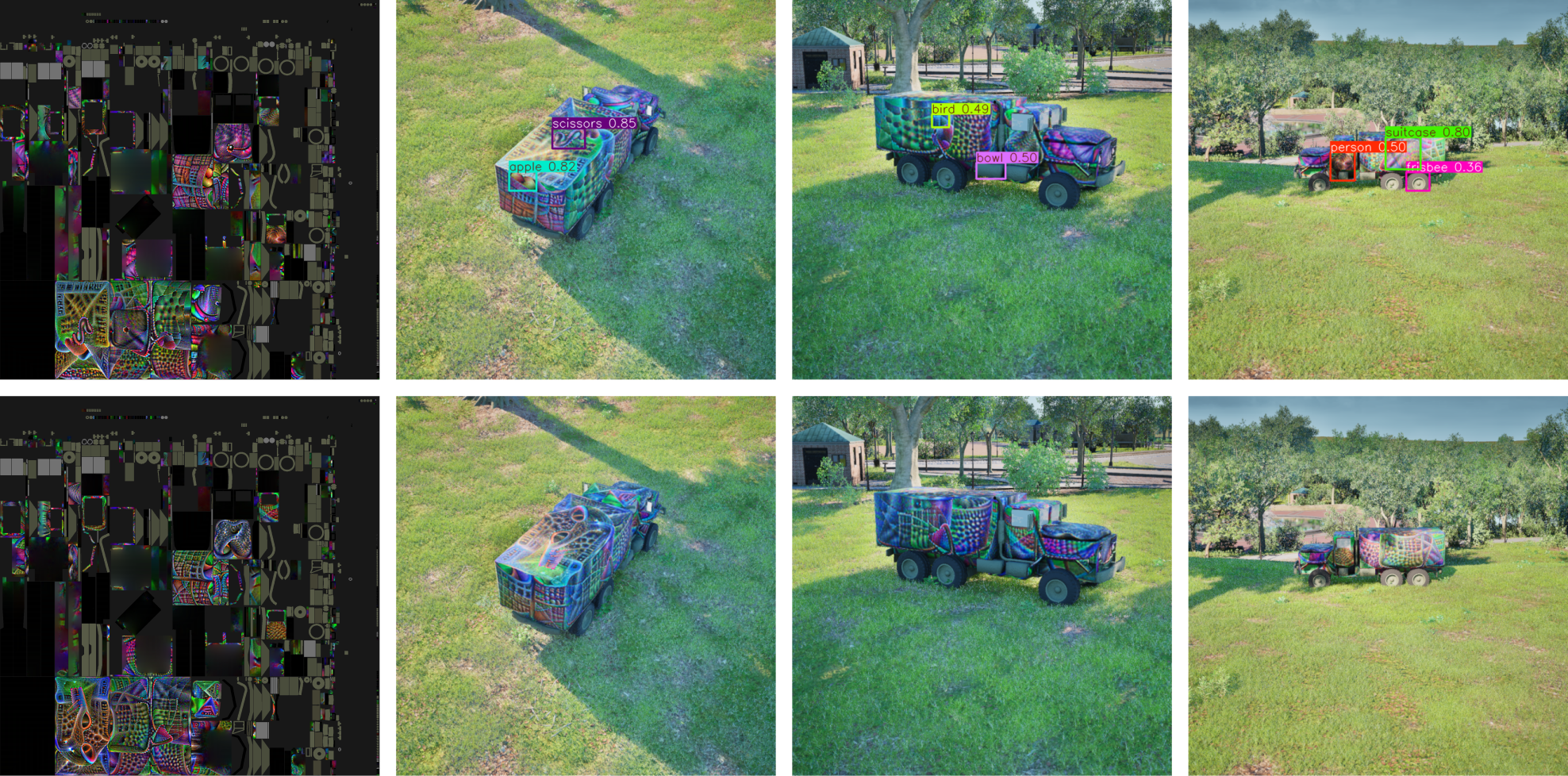}
    \caption{Comparison 
 of IoU-based and IoP-based bounding box filtering for the class loss. Top row: IoU-based filtering results in false-positive detections on the truck surface. Bottom row: IoP-based filtering suppresses these false positives. Each column shows a different viewpoint of the same optimized~texture.}
    \label{fig:iop_vs_iou_filtering}
\end{figure}

{
To address this, we developed IoP-based filtering, which prioritizes bounding boxes based on the proportion of their area overlapping with the truck, ensuring that all relevant regions are included in the optimization. By~capturing these inner regions, IoP-based filtering prevents the generation of patterns that trigger false detections. Figure~\ref{fig:iop_vs_iou_filtering} illustrates this difference; IoU-based filtering results in adversarial patterns that produce false positives across the truck’s surface, while IoP-based filtering eliminates such detections, rendering the truck undetectable.
}

{
In addition to the class loss with IoP-based filtering, we incorporate an IoU loss term to further suppress the detector's ability to localize the truck accurately. 
}
The IoU loss is defined as
\begin{equation} 
\mathcal{L}_{IoU} = \sum_{i \in \Omega_{IoU}} \text{IoU}(B_{pred}^i, B_{gt})
\end{equation}
where:
\begin{itemize}
    \item $\Omega_{IoU} = \{ i \mid \text{IoU}(B_{pred}^i, B_{gt}) > \tau_{IoU} \}$ denotes 
 the set of predicted bounding boxes with relatively large IoU values that we want to minimize.\textbf{}
\end{itemize}

Finally, the~total attack loss is the weighted sum of the class confidence loss and the IoU loss, with~a weight $\beta$ to balance the contribution of the IoU loss term:
\begin{equation} \mathcal{L}_{atk} = \mathcal{L}_{cls} + \beta \mathcal{L}_{IoU} \end{equation}
By minimizing this loss, we effectively reduce the confidence scores for any objects detected over the area of our truck.

\subsubsection{Convolutional Smooth~Loss}
\label{sec:smooth_loss}
An additional constraint is to generate physically producible adversarial patterns by ensuring a smooth texture. The~traditional smoothness loss function calculates the Total Variation (TV) between adjacent pixels, specifically looking at the immediate right and bottom neighbors~\cite{TV_loss}:
\begin{equation}
\mathcal{L}_{\text{TV}} = \sum_{i,j} \sqrt{(\delta_{i,j} - \delta_{i+1,j})^2 + (\delta_{i,j} - \delta_{i,j+1})^2}
\end{equation}

To improve upon this, we introduce Convolutional Smooth Loss—a generalized approach which is the extension of the traditional TV loss. Rather than only considering the immediate right and bottom neighbors, this method evaluates the differences between the central pixel and all pixels within a $k \times k$ kernel. This captures the local variation over a larger neighborhood. Mathematically, let $T_{i,j}$ represent the pixel value at position $(i,j)$. We calculate the local variation $D_{i,j}$ as the sum of squared differences between $T_{i,j}$ and all pixels within the $k \times k$ neighborhood:
\begin{equation}
D_{i,j} =  \sum_{n=-\lfloor \frac{k}{2} \rfloor}^{\lfloor \frac{k}{2} \rfloor} \sum_{m=-\lfloor \frac{k}{2} \rfloor}^{\lfloor \frac{k}{2} \rfloor}(T_{i,j} - T_{i+n,j+m})^2
\end{equation}

The overall smoothness loss is then computed as
\begin{equation}
\mathcal{L}_{\text{smooth}}(T) = \frac{1}{W\cdot H}\sum_{i,j}\sqrt{D_{i,j}}
\end{equation}
where $W$ and $H$ are the width and height of the texture image respectively. Note that the term $D_{i,j}$ can be easily calculated for every pixel of the image with the use of two convolutions. Let $*$ denote the convolution operation, and~$K$ a $k\times k$ kernel with weights $K_{n,m} = \frac{1}{k^2}$ (assuming a uniform weighting for simplicity). Using this kernel, we can compute $D$ for the entire image in a compact and efficient manner as
\begin{equation}
    D = k^2T^2 - 2T \cdot (T * K) + T^2 * K
\end{equation}
where $T^2$ refers to the element-wise square of the pixel values. $T*K$ is the convolution of the texture $T$ with the kernel $K$, while $T^2*K$ is $T^2$ convoluted with $K$.
This formulation not only captures smoother transitions over a larger neighborhood but also allows for fast computation with convolutions. The~ability to choose $k$ arbitrarily gives us further control over the degree of~smoothness.

Finally, the~total loss function used in our optimization process combines the adversarial attack loss $\mathcal{L}_{atk}$ with the smoothness loss $\mathcal{L}_{\text{smooth}}$, weighted by a factor $\gamma$:
\begin{equation}
    \mathcal{L}_{total} = \mathcal{L}_{atk} + \gamma \mathcal{L}_{smooth}
\end{equation}

\subsubsection{Projected Gradient Descent with~Adam}
\label{sec:pgd}

When optimizing the adversarial texture \( T_{adv} \), each pixel must remain within the valid range \([0,1]\). 
{
\textls[-15]{Formally, this can be viewed as the following constrained optimization~problem:}
}
\begin{equation}
\min_{T_{adv} \in [0, 1]^{3 \times H \times W}} \,\mathcal{L}(T_{adv})    
\end{equation}
where \(\mathcal{L}\) is our total loss function 
{
(combining both adversarial and smoothness terms). A~standard approach to such a box-constrained problem is Projected Gradient Descent (PGD), which clips parameter values back into the feasible range after each update. PGD-based methods have become standard in adversarial machine learning because they naturally ensure the generated perturbations satisfy domain-specific constraints (e.g., \(\ell_p\)-norm bounds or pixel-value ranges~\cite{pgd}).
}

\paragraph{Standard Clipping and Its Limitations}
{
A common strategy enforces \([0,1]\) constraints by simply clipping the texture values after each update step. While this guarantees feasibility, it can “zero out” critical gradient information if many pixels are pushed outside the valid range, slowing convergence and yielding suboptimal solutions.
}
\paragraph{PGD with Adam}

{
To overcome these issues, we combine PGD with the Adam~\cite{adam} optimizer. Adam adaptively scales the learning rate per parameter and typically converges faster in high-dimensional tasks such as texture optimization. In~this context, we use the notation \(\nabla_{\text{Adam}}\mathcal{L}\) to denote the gradient update calculated by the Adam optimizer. A~naive combination of Adam with PGD would follow this sequence:
}
\begin{enumerate}
    \item Compute Raw Updates: 
    \[
    \Delta T = \nabla_{\text{Adam}}\mathcal{L}({T_{adv}^t}),
    \]
    where \(\Delta T\) is the parameter update vector derived from~Adam.

    \item Update step: 
    \[
    T_{adv}^{t+1} \leftarrow T_{adv}^{t} - \eta\ \Delta T
    .\]
    \item Texture projection step: 
    \[
    T_{adv}^{t+1} \leftarrow \min\Bigl(\,\max\bigl(\,T_{adv}^{t+1}, \, 0 \,\bigr),\,1 \Bigr).
    \]
\end{enumerate}
{
While this approach guarantees feasibility, it disrupts gradient information whenever pixel values are pushed outside the valid range, resulting in less efficient optimization.
}
\paragraph{Gradient Projection}
{
Instead of clipping the texture values after the update, we propose modifying the gradients before applying them. This approach avoids the loss of gradient information while also maintaining feasibility:
}
\begin{enumerate}
    \item Compute Raw Updates: 
    \[
    \Delta T = \nabla_{\text{Adam}}\mathcal{L}({T_{adv}^t}).
    \]

    \item Gradient projection step: 
    \[
    \nabla_{\text{proj}} \mathcal{L}(T_{adv}^t) 
      = \min\Bigl(\,\max\bigl(\Delta T,\, -\,T_{adv}^t\bigr),\,1 \,-\,T_{adv}^t\Bigr).
    \]

    \item Update step: 
    \[
    T_{adv}^{t+1} = T_{adv}^t - \eta\, \nabla_{\text{proj}} \mathcal{L}(T_{adv}^t).
    \]
\end{enumerate}

{
This gradient projection approach preserves gradient flow, leading to more stable and efficient optimization compared to naive clipping. The~overall adversarial texture optimization process is illustrated in Figure~\ref{fig:texture_optimization}, showing how the Neural Renderer, Object Detector, and~Gradient Projection blocks work together to iteratively update the texture.
}

\section{Experimental~Setup}\label{section4}
\unskip

\subsection{Truck~Model}
\label{sec:truck_model}
\begin{figure}[H]
    
    \includegraphics[width = 0.85\textwidth]{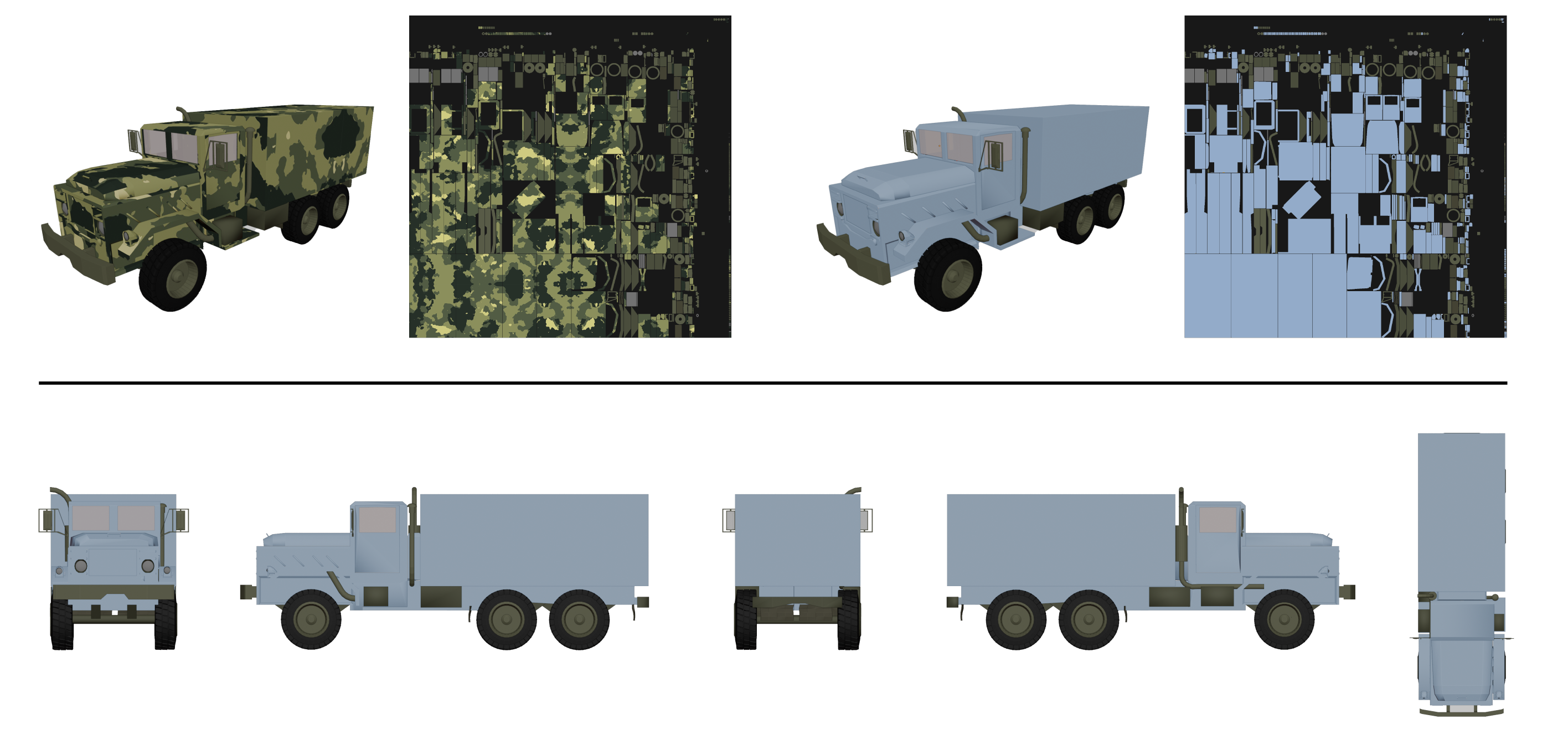}
    \caption{The 3D truck model with the corresponding 2D textures. Top left: a naive, simple camouflage. Top right: blue paint indicates body parts that are used for the adversarial attack. The~bottom row displays various views of the truck, showing the parts involved in the adversarial~attack.}
    \label{fig:truck_model}
\end{figure}

In our framework, we utilized an M923 truck model consisting of 24,784 faces (Figure~\ref{fig:truck_model}). To~improve computational efficiency during adversarial texture generation, we divided the truck into two distinct parts based on which components are typically painted in real-life~scenarios:

\begin{itemize} 
    \item \textbf{Body Parts:} This segment includes the main body of the truck, such as the carrosserie and tarp, which are usually painted for camouflage purposes. While these body parts comprise only 1282 triangular faces (~5\% of the total number of faces), they cover approximately 80\% of the truck's visible surface area. This is due to the fact that these regions consist of larger, less intricate surfaces compared to the auxiliary parts like~wheels.
    \item \textbf{Auxiliary Parts:} The remaining components of the truck—such as the wheels, bumper, exhaust stacks, and~other parts that are not typically painted or are impractical to paint—fall under this category. 
    {
    These parts contain the remaining 23,502 triangular faces (~95\% of the faces), but~they account for only ~20\% of the truck’s visible surface area due to their smaller individual sizes and more intricate geometry.
    }
\end{itemize}

By separating the truck into these two parts, we ensure that only the body part (approximately 5\% of the full truck model) is processed through the neural renderer. This not only reduces the computational load but also significantly lowers the memory requirements for the differential renderer, making the overall pipeline more~efficient.

It is important to note that the full truck model is rendered within the UE5 environment. Because~of this, the~auxiliary parts of the vehicle are still visible in the final image, as~they are blended with the background using the binary mask $M$ applied to the reference image $X_{ref}$. This allows us to focus the adversarial attack exclusively on the parts of the truck that could realistically be~modified.

\subsection{Dataset} 
\label{sec:dataset} 
{To develop and evaluate our adversarial camouflage approach, we constructed a custom dataset in Unreal Engine 5 (UE5) featuring an M923 3D truck model. This dataset plays a dual role; it is used both for training the PRN and for guiding the adversarial texture optimization. We organized the data generation process into two main~parts: 
\begin{itemize}
    \item \textbf{Core Truck Dataset:} A large collection of rendered truck images under diverse positions, camera parameters, and~textures. 
    \item \textbf{High-Resolution Texture Dataset:} A complementary set of 4,000 high-resolution texture images. 
\end{itemize}
}
\subsubsection{Core Truck~Dataset}
{We first identified 25 distinct truck locations within the UE5 scene. At~each location, we placed the M923 truck and generated 2000 rendered images by randomly sampling the~following: 
\begin{itemize} 
    \item Camera viewpoints and distances: For each image, the~camera was placed between 5~m and 35~m from the truck, randomly oriented with azimuth angles between 5$^\circ$ and 90$^\circ$, and~elevation angles from 0$^\circ$ to 360$^\circ$.
    \item \textls[-21]{Truck textures: A random selection from our High-Resolution Texture Dataset (see~below).} 
\end{itemize} 
By combining 25 locations with 2,000 images each, we obtained a total of 50,000 images. We refer to this entire set as the \textbf{Core Truck Dataset}. Along with each rendered image $X_{ref}$, we also stored the gray textured image $X_{gray}$, depth map $X_d$, binary mask $M$, camera parameters $\theta_c$, and~the chosen texture $T$. This dataset size was chosen incrementally. Smaller datasets tested during preliminary experiments (e.g., 200 images per position) often led to poor generalization of the neural renderer, with~artifacts appearing on unseen textures or views. Incrementally increasing the dataset size—from a few thousand images to tens of thousands—significantly improved the PRN’s ability to render previously unseen views and texture patterns accurately.
}

\subsubsection{High-Resolution Texture~Dataset}
{To produce visually diverse truck appearances, we built a texture library of 4,000~images at $2048\times2048$ resolution. The~library includes the following: 
}
\begin{itemize} 
    \item \textbf{Describable Textures Dataset} \cite{DTD}: A total of 1500 images were carefully selected from this dataset to provide a variety of texture patterns. 
    \item \textbf{Van Gogh Paintings} \cite{VGP}: A total of 300 texture images were sourced from Van Gogh’s paintings, chosen for their distinct color patterns. 
    \item \textbf{Random Uniform Color Images}: A total of 200 images consisting of uniform color values were generated. 
    \item \textbf{Random Noise Images}: A total of 2000 noise textures were randomly generated to simulate non-structured patterns visually similar to adversarial patterns. 
\end{itemize}

\begin{figure}[H]
\begin{adjustwidth}{-\extralength}{0cm}
    \centering
    \includegraphics[width=\linewidth]{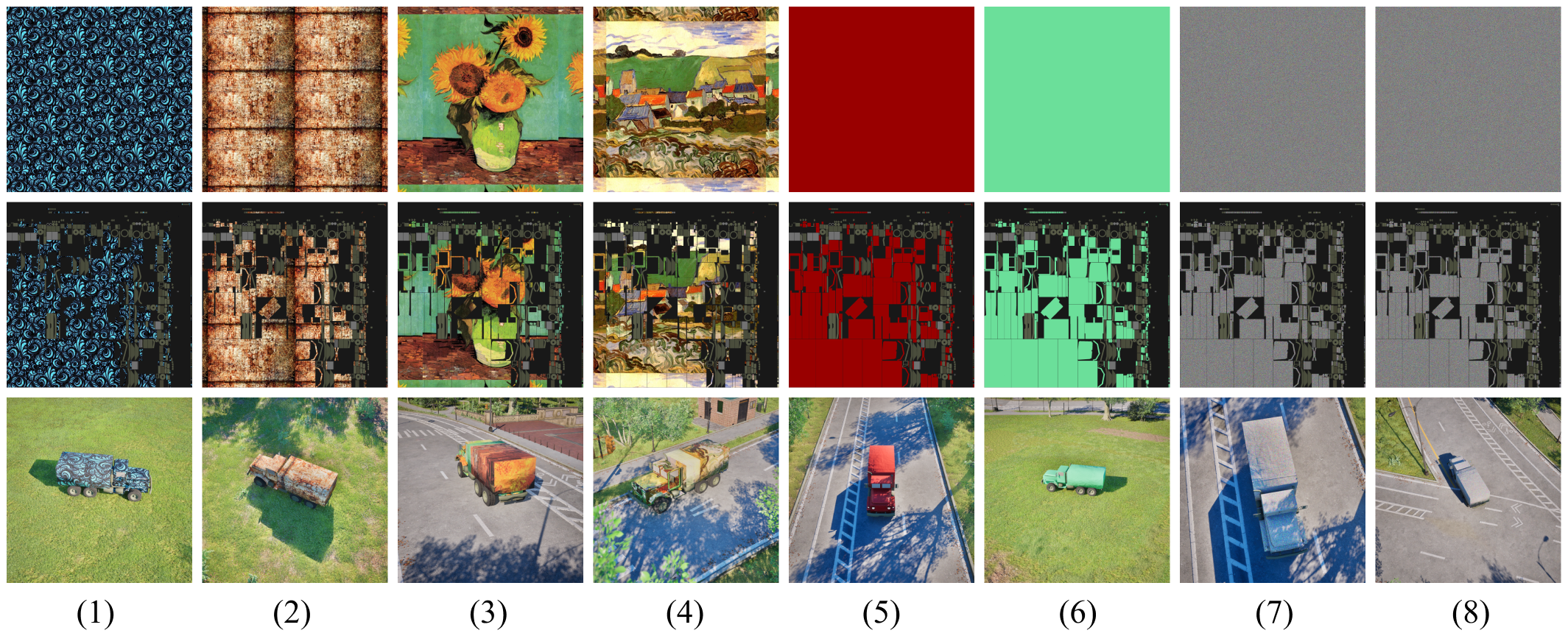}
    \end{adjustwidth}
    \caption{Examples 
 from the High-Resolution Texture Dataset. Each column showcases textures from one of the four categories in the dataset: (1--2) Describable Textures Dataset, (3--4) Van Gogh Paintings, (5--6) Random Uniform Color Images, and~(7--8) Random Noise Images. The~first row presents the full texture images, the~second row shows masked versions where only the truck’s body parts are textured (as described earlier in Section~\ref{sec:truck_model}), and~the third row displays the textures rendered on the truck in a specific scene from the Core Truck~Dataset.}
    \label{fig:HRTD}

\end{figure}

{
This broad collection helps our neural renderer generalize effectively to different surface appearances. To~provide a visual overview of the High-Resolution Texture Dataset, Figure \ref{fig:HRTD} 
 presents representative samples from each of the four texture categories. The~first row displays the original full-resolution textures. The~second row demonstrates the application of a masking process to focus the texture exclusively on the truck’s body parts, blending it with a base texture for auxiliary components like wheels. Finally, the~third row illustrates how these textures appear when rendered on the truck in a specific scene, as~part of the Core Truck Dataset. 
}

\subsection{Implementation~Details}
\unskip

\subsubsection{PRN~Training}
\label{sec:prn_training}
{For training the PRN, we used the entire 50,000 image Core Truck Dataset. To~ensure that the PRN can handle unseen textures, we split the 4000 texture library into 3250 textures for training and 750 textures for testing. Consequently, images in the Core Truck Dataset whose textures came from the 3250-texture subset were grouped as the PRN training set 
, and~images whose textures came from the 750-texture subset served as the PRN test set. This approach yields a total of 40,625 training images and 9375 testing images. We used the Adam optimizer with a learning rate of 0.0001 and trained it for 100 epochs. The~final model achieved an L1 rendering loss of 0.024 and an SSIM of 0.9901 on the validation set.
}

\subsubsection{Adversarial Texture~Generation} 
{While the PRN training used the entire Core Truck Dataset, our adversarial texture optimization only requires a subset. Specifically, in this case, we performed the following: 
\begin{itemize} 
    \item We used 1000 images per truck location (instead of the full 2000) to reduce training time. This subset thus contains 25,000 images in total. 
    \item We performed a location-based split into 18 locations for training (18,000 images) and 7 locations for testing (7000 images).
\end{itemize}
This ensures that during the adversarial optimization phase, the~model sees a wide range of camera perspectives and scenes in the training split, while the evaluation is conducted on entirely different locations of the truck, enforcing a strict generalization test. For every training iteration, the~truck’s original rendered texture is dynamically replaced by the adversarial texture under optimization 
 $T_{adv}$. The~neural renderer thus re-synthesizes the scene with the latest version of the adversarial texture, and~we compute gradients (via our loss functions) to update the texture’s pixel values. We used the Adam optimizer with gradient projection as described in Section~\ref{sec:pgd} with a learning rate of 0.006 over 6 epochs on the 18,000 training samples.}

We set the IoU loss term weight to $\beta = 0.01$ and the smoothness regularization term weight to $\gamma = 0.1$ with a convolution kernel size of $k = 3$ for this configuration. The~IoP threshold was set to $\tau_{IoP} = 0.6$ and the IoU threshold to $\tau_{IoU} = 0.45$. These settings were used as the default configuration in all subsequent~experiments.

In evaluations where we tested different setups or other hyperparameters, the~remaining parameters (such as the learning rate, the~initialization, and~the loss weights) were kept consistent with the default configuration unless otherwise~noted.

\section{Results}\label{section5}

In this section, we evaluate the effectiveness of the proposed TACO framework. We conduct a series of experiments to assess the attack performance against various object detection models, analyse the impact of different loss components, examine the influence of texture initialization strategies, and~visualize the attention shifts in the target model using Class Activation Maps (CAMs). 

\subsection{Evaluation Metrics and~Models} 
\label{sec:evaluation_models}
All evaluations are conducted on the test set of eight unseen truck positions. Since trucks can sometimes be mistaken for cars by object detection models due to their similar appearance, we treat cars and trucks as a single combined class when calculating our metrics. We adopt two primary metrics to assess the performance of our adversarial~textures:

\begin{itemize} 
    \item Average Precision at IoU threshold 0.5 (\textbf{AP@0.5}): This metric evaluates the precision of the object detector when the Intersection over Union (IoU) between the predicted bounding box and the ground truth exceeds 50\%. 
    \item Attack Detection Rate (\textbf{ADR}): The proportion of images in which the object detector successfully identifies the truck. \end{itemize}

We evaluate our adversarial textures against six object detection models:

\begin{itemize} 
    \item \textbf{YOLOv8X} \cite{yolov8_ultralytics}: Our target model for the adversarial attack. 
    \item \textbf{YOLOv3u} \cite{yolov3}: A previous generation of the YOLO family with an upgraded detection head. 
    \item \textbf{YOLOv5Xu}: \textls[-25]{An intermediate version of the YOLO models with upgraded detection~heads. }
    \item Faster R-CNN v2 (\textbf{FRCNN}) \cite{FRCNNv2}: An improved version of the two-stage object detection model. 
    \item Fully Convolutional One-Stage Object Detection (\textbf{FCOS}) \cite{FCOS}: An anchor-free object detection framework. 
    \item Detection Transformer (\textbf{DETR}) \cite{DETR}: A transformer-based object detection model. 
\end{itemize}

All models are pre-trained on the COCO dataset~\cite{COCO} and are treated as black-box models except for YOLOv8, our white-box~target.

\subsection{Performance Comparison of Different~Textures} 

We first compare the effectiveness of our TACO-generated adversarial texture against three baseline textures:

\begin{itemize} 
    \item \textbf{Base}: \textls[-15]{The original single-color texture of the truck without any adversarial modifications.} 
    \item \textbf{Naive}: A simple camouflage texture common on military trucks. 
    \item \textbf{Random}: A texture initialized with random pixel values.\item {
    \textbf{DTA}: The Differentiable Transformation Network Approach (DTA), an~existing adversarial camouflage method reimplemented for comparison~\cite{DTA}.
    }
\end{itemize}

\begin{figure}[H]
    \centering
    \includegraphics[width=\textwidth]{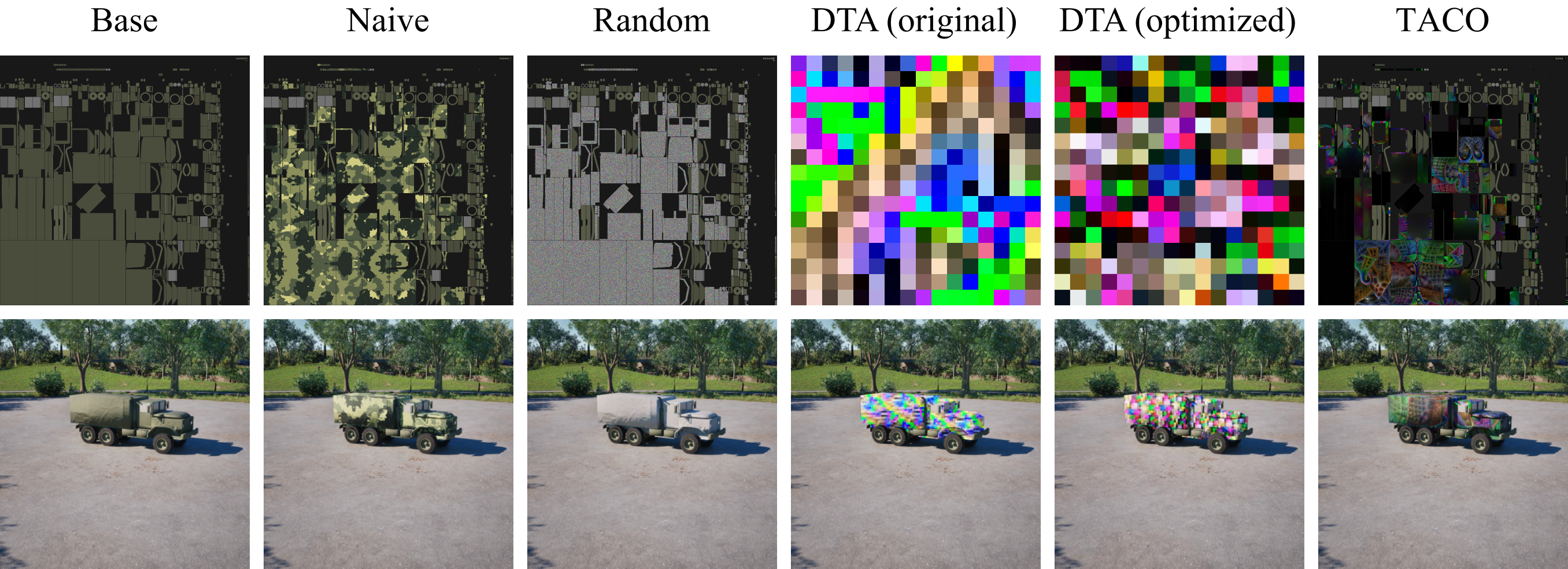}
    \caption{Visual 
 comparison of the Base, Naive, Random, DTA, and~TACO textures (first row) and their application to the truck model (second row).}
    \label{fig:textures}
\end{figure}

Figure~\ref{fig:textures} shows the applied textures for the Base, Naive, Random, DTA, and~TACO methods used in this evaluation. 
{
In the case of DTA, we include two variants of the texture: one directly from the original paper (DTA (original)) and another optimized specifically for our truck model and environment (DTA (optimized)). For~fairness, only the re-optimized DTA texture is used in our experiments. Additionally, both DTA textures are rendered photorealistically using our own PRN.
} Tables~\ref{tab:textures_ap50} and ~\ref{tab:textures_adr} present the AP@0.5 and ADR results, respectively, across all evaluated~models.

\vspace{-8pt}

\begin{figure}[H] 
 
\includegraphics[width=0.85\textwidth]{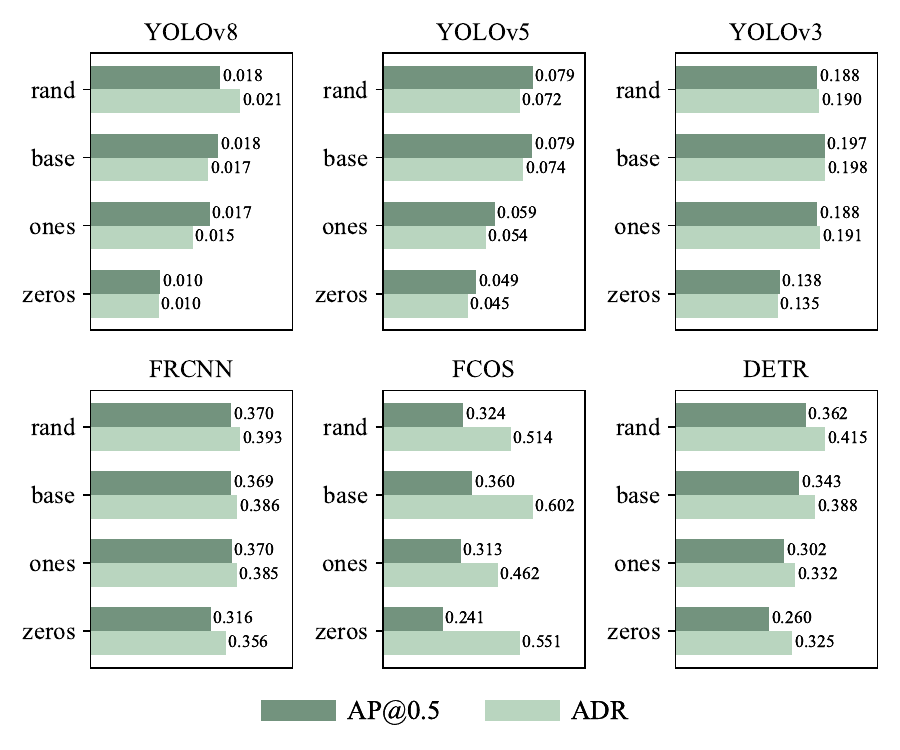} 
\caption{Impact of different texture initialization methods on attack~performance.} 
\label{fig:column_diagram} 
\end{figure}

The texture generated by TACO significantly reduces both the AP@0.5 and ADR across all evaluated models compared to the baseline textures. Specifically, for~YOLOv8, the~AP@0.5 drops from 0.7295 (Base) to 0.0099 (TACO), and~the ADR decreases from 0.7453 to 0.0097. This indicates that the adversarial texture effectively deceives the target model, rendering the truck nearly undetectable in unseen scenes.
While TACO is optimized specifically against YOLOv8, it also exhibits transferability to other models, although~with varying degrees of effectiveness. This suggests that our adversarial texture exploits common vulnerabilities across different object detection~architectures.

\begin{table}[H]

\caption{AP@0.5 performance comparison of different textures across various object detection~models.}
\begin{tabularx}{\textwidth}{lCCCCCC}
\toprule
\multirow{2}{*}{\vspace{-6pt}\textbf{Method}} & \multicolumn{6}{c}{\textbf{AP@0.5}} \\ 
\cmidrule{2-7} 
\addlinespace[1mm] 
& \textbf{YOLOv8} & \textbf{YOLOv3} & \textbf{YOLOv5} & \textbf{FRCNN} & \textbf{FCOS} & \textbf{DETR} \\
\midrule
Base & 0.7295 & 0.7216 & 0.6132 & 0.8377 & 0.6361 & 0.6441 \\
Naive & 0.8057 & 0.7305 & 0.6518 & 0.7770 & 0.6317 & 0.6619 \\
Random & 0.6705 & 0.7214 & 0.5537 & 0.8202 & 0.5948 & 0.6470\\
DTA (optimized) & 0.2865 & 0.3663 & 0.3068 & 0.4532 & 0.3633 & 0.4121\\
TACO & \textbf{0.0099} & \textbf{0.0491} & \textbf{0.1381} & \textbf{0.3157} & \textbf{0.2410} & \textbf{0.2600} \\
\bottomrule
\end{tabularx}
\label{tab:textures_ap50}
\end{table}
\unskip

\begin{table}[H]

\caption{ADR performance comparison of different textures across various object detection~models.}
\begin{tabularx}{\textwidth}{lCCCCCC}
\toprule
\multirow{2}{*}{\vspace{-6pt}\textbf{Method}} & \multicolumn{6}{c}{\textbf{ADR}} \\ 
\cmidrule{2-7} 
\addlinespace[1mm] 
& \textbf{YOLOv8} & \textbf{YOLOv3} & \textbf{YOLOv5} & \textbf{FRCNN} & \textbf{FCOS} & \textbf{DETR} \\
\midrule
Base & 0.7453 & 0.7258 & 0.6195 & 0.8689 & 0.7475 & 0.6678 \\
Naive & 0.8241 & 0.7376 & 0.6561 & 0.8234 & 0.7719 & 0.7045 \\
Random & 0.6814 & 0.7313 & 0.5614 & 0.8440 & 0.6959 & 0.6586 \\
DTA (optimized) & 0.2906 & 0.3646 & 0.3011 & 0.4560 & \textbf{0.3851} & 0.4130\\
TACO & \textbf{0.0097} & \textbf{0.0448} & \textbf{0.1354} & \textbf{0.3558} & 0.5511 & \textbf{0.3247} \\
\bottomrule
\end{tabularx}
\label{tab:textures_adr}
\end{table}
\unskip

\subsection{Impact of Different Loss~Functions} 

We investigate the contribution of each component in our total loss function by evaluating the adversarial textures generated using different combinations of the proposed loss terms. Tables~\ref{tab:losses_ap50} and~\ref{tab:losses_adr} present the AP@0.5 and ADR results, respectively, for~each loss configuration including $\mathcal{L}_{cls}$, $\mathcal{L}_{cls} + \mathcal{L}_{iou}$, $\mathcal{L}_{cls} + \mathcal{L}_{sm}$, and~the full $\mathcal{L}_{total}$ across all~models.

\begin{table}[H]

\caption{AP@0.5 performance comparison of different loss~schemes.}
\begin{tabularx}{\textwidth}{lCCCCCC}
\toprule
\multirow{2}{*}{\vspace{-6pt}\textbf{Method}} & \multicolumn{6}{c}{\textbf{AP@0.5}} \\ 
\cmidrule{2-7} 
\addlinespace[1mm] 
& \textbf{YOLOv8} & \textbf{YOLOv3} & \textbf{YOLOv5} & \textbf{FRCNN} & \textbf{FCOS} & \textbf{DETR} \\
\midrule
$\mathcal{L}_{cls}$ & 0.0283 & 0.1283 & 0.1976 & 0.4884 & 0.4400 & 0.4280 \\
$\mathcal{L}_{cls} + \mathcal{L}_{iou}$ & 0.0189 & 0.0690 & 0.1677 & 0.3786 & 0.2586 & 0.3105 \\
$\mathcal{L}_{cls} + \mathcal{L}_{sm}$& 0.0373 & 0.1283 & 0.2371 & 0.4979 & 0.4892 & 0.4638\\
$\mathcal{L}_{total}$ & \textbf{0.0099} & \textbf{0.0491} & \textbf{0.1381} & \textbf{0.3157} & \textbf{0.2410} & \textbf{0.2600}
 \\
\bottomrule
\end{tabularx}
\label{tab:losses_ap50}
\end{table}
\unskip

\begin{table}[H]

\caption{ADR performance comparison of different loss~schemes.}
\begin{tabularx}{\textwidth}{lCCCCCC}
\toprule
\multirow{2}{*}{\vspace{-6pt}\textbf{Method}} & \multicolumn{6}{c}{\textbf{ADR}} \\ 
\cmidrule{2-7} 
\addlinespace[1mm] 
& \textbf{YOLOv8} & \textbf{YOLOv3} & \textbf{YOLOv5} & \textbf{FRCNN} & \textbf{FCOS} & \textbf{DETR} \\
\midrule
$\mathcal{L}_{cls}$ & 0.0317 & 0.1262 & 0.1953 & 0.5067 & 0.5542 & 0.4451\\
$\mathcal{L}_{cls} + \mathcal{L}_{iou}$ & 0.0189 & 0.0690 & 0.1677 & 0.3786 & \textbf{0.2586} & \textbf{0.3105} \\
$\mathcal{L}_{cls} + \mathcal{L}_{sm}$& 0.0361 & 0.1298 & 0.2317 & 0.5097 & 0.6220 & 0.4867\\
$\mathcal{L}_{total}$ & \textbf{0.0097} & \textbf{0.0448} & \textbf{0.1354} & \textbf{0.3558} & 0.5511 & 0.3247\\
\bottomrule
\end{tabularx}
\label{tab:losses_adr}
\end{table}

Optimizing with only the classification loss ($\mathcal{L}_{cls}$) reduces the detection performance, but~not as significantly as when additional loss terms are included. Incorporating the IoU loss ($\mathcal{L}_{iou}$) further decreases both AP@0.5 and ADR, particularly for models like FCOS and DETR, indicating that minimizing bounding box overlap enhances the attack's~effectiveness.

Adding the smoothness loss ($\mathcal{L}_{sm}$) slightly reduces the attack performance compared to using $\mathcal{L}_{cls}$ alone. However, it contributes to generating more even textures, which is important for real-world~applicability.

The full loss function ($\mathcal{L}_{total}$) achieves the best overall performance on YOLOv8 and maintains competitive results on other models. This demonstrates that combining all loss components effectively contributes to attack strength and texture~plausibility.

\subsection{Texture Initialization~Study} 

We examine the impact of different texture initialization strategies on the optimization outcome. The~initialization methods tested are as follows:

\begin{itemize} 
    \item \textbf{Zeros}: Initializing the texture with all zeros (black texture). 
    \item \textbf{Ones}: Initializing the texture with all ones (white texture). 
    \item \textbf{Random}: Initializing with random values. 
    \item \textbf{Base}: Starting from the truck's original texture. \end{itemize}

Figure~\ref{fig:column_diagram} presents a column diagram comparing the AP@0.5 and ADR for each initialization method across all~models.
\vspace{-3pt}

\begin{figure}[H] 
 
\includegraphics[width=\linewidth]{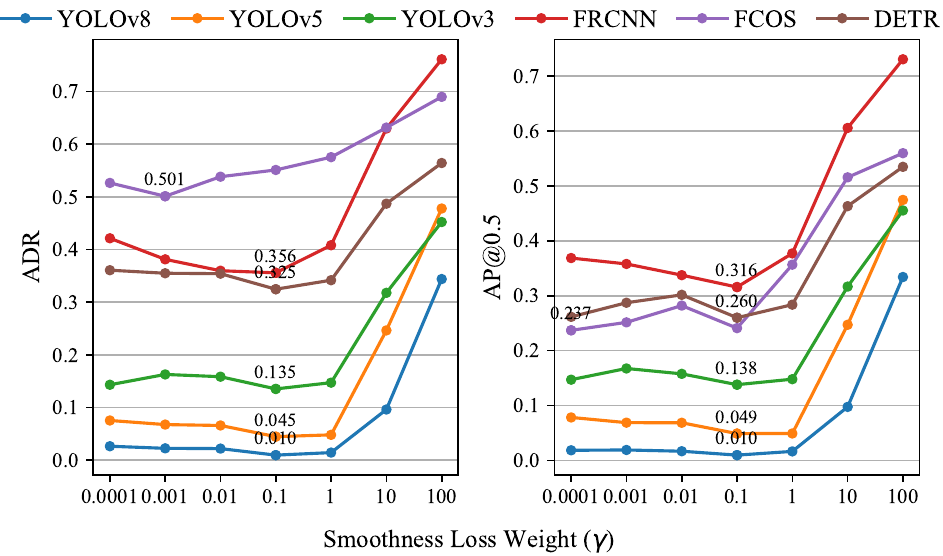} \caption{Effect of varying the smoothness loss coefficient ($\gamma$) on attack~performance.} 
\label{fig:smooth_loss_coeff} 
\end{figure}

Initializing the texture from zeros yields the best adversarial performance in terms of both AP@0.5 and ADR across most models, except~for FCOS, where initializing from ones performs slightly better in terms of ADR. Initializing from the base texture or random values results in less effective~attacks. 

\subsection{Smoothness Loss Coefficient~Analysis} 

We explore how varying the smoothness loss coefficient ($\gamma$) affects the adversarial texture's effectiveness and visual quality. Figure~\ref{fig:smooth_loss_coeff} illustrates the relationship between different $\gamma$ values and the corresponding AP@0.5 and ADR on the object~detectors.

When $\gamma > 1$, the~smoothness loss dominates the optimization, resulting in overly smooth textures that lack the necessary perturbations to deceive the detector, thereby reducing the attack's effectiveness. Conversely, when $\gamma$ is too small (e.g., $\gamma < 0.01$), the~texture may become overly noisy, potentially making it visually~unrealistic.

\begin{figure}[H]
\begin{adjustwidth}{-\extralength}{0cm}
    \centering
    \includegraphics[width=\linewidth]{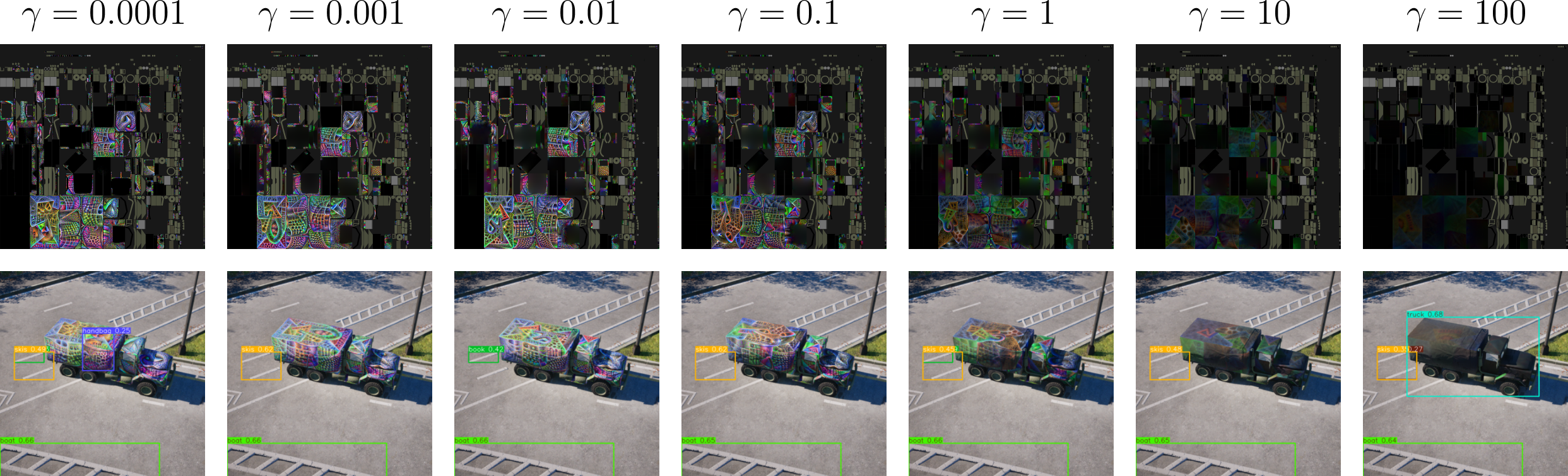}
    \end{adjustwidth}
    \caption{Textures 
 generated with varying smoothness loss coefficients ($\gamma$) and their application to the truck~model.}
\label{fig:textures_trucks}

\end{figure}

In addition to the performance metrics, Figure~\ref{fig:textures_trucks} 
 provides a visual representation of how the generated textures evolve with different $\gamma$ values and~how these textures appear when applied to the truck. The~second row of the figure shows the truck with the textures applied, along with the YOLOv8 detection results. As~$\gamma$ increases, the~textures become smoother. When $\gamma$ reaches 100, the~texture becomes overly smooth and resembles a plain black color, which diminishes its adversarial~effect.

 We observe a sweet spot in the range $0.01 < \gamma < 1$, where the attack achieves optimal performance while maintaining a balance between texture smoothness and adversarial perturbations. A~$\gamma$ value around 0.1 provides a good trade-off, leading to effective attacks across all~models.

Interestingly, the~adversarial perturbations not only cause the detector to miss the truck but also lead to hallucinations of other objects outside the truck. As~shown in Figure~\ref{fig:textures_trucks}, YOLOv8 detects phantom objects in the surrounding areas where there are none. This phenomenon occurs because object detection is not a task that is separable from its surrounding context. As~demonstrated in~\cite{background_attack}, even perturbing the background of an object can significantly affect detection performance. By~altering the texture of the truck, the~attack can indirectly affect how the model interprets other parts of the~scene.

\subsection{Class Activation Mapping~Analysis} 
\begin{figure}[H] 
 
\includegraphics[width=0.7\textwidth]{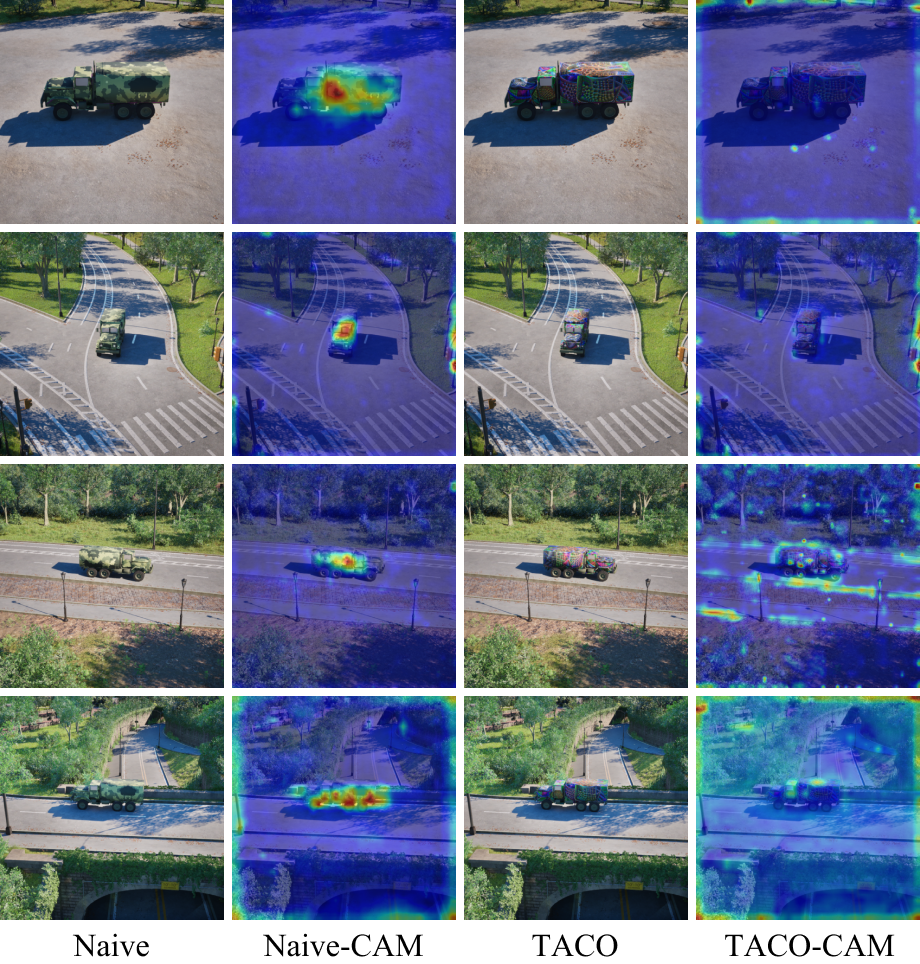} 
\caption{Ablation 
 CAM for YOLOv8; the heatmaps indicate the regions where the model focuses its~attention.} \label{fig:CAM} 
\end{figure}

To understand how the adversarial texture affects the target model's attention mechanisms, we adapt Ablation-CAM~\cite{Ablation_CAM} to visualize the regions of interest in YOLOv8. Figure~\ref{fig:CAM} displays the CAM overlays on the truck images with and without the adversarial~texture.

In the original image without the adversarial texture, the~CAM highlights the truck, indicating that YOLOv8 correctly focuses on the vehicle for detection. After~applying the TACO adversarial texture, the~CAM shifts away from the truck, and~the model's attention is dispersed to surrounding areas. This suggests that the adversarial texture successfully misleads the model's attention mechanisms, contributing to the failure in detecting the~truck.

\section{Conclusions}\label{section6}
In this paper, we introduced TACO (Truck Adversarial Camouflage Optimization), a~novel framework for generating adversarial camouflage patterns on 3D vehicle models to deceive state-of-the-art object detection systems. By~using UE5 for photorealistic rendering and a novel neural renderer component, TACO optimizes textures that are both visually smooth and highly effective in deceiving detectors. We introduced the Convolutional Smoothness Loss, which ensures that the generated patterns maintain a realistic appearance. Experimental results showed that our adversarial textures significantly reduced the detection performance of YOLOv8, achieving a near-zero AP@0.5 and ADR on unseen test data. The~adversarial patterns also exhibited transferability to other object detection models including other YOLO versions, Faster R-CNN, and~DETR. We also showed that balancing texture smoothness with adversarial perturbations is crucial for optimal performance. By~targeting YOLOv8, we advanced the field beyond previous works that focused on older detection models, demonstrating the viability of adversarial attacks against more robust and modern~architectures.

{
Despite these promising results, our approach has areas for further exploration. First, we tested it on a single truck model, leaving room to evaluate its effectiveness across diverse vehicle models. Such an extension would provide a more comprehensive assessment. Second, while TACO demonstrates strong transferability, the~training pipeline and textures are primarily optimized for YOLO-based architectures. Investigating domain adaptation or multi-objective optimization methods could expand its applicability to a broader range of detectors.
Additionally, our dataset, though~leveraging UE5 for photorealistic rendering, does not yet account for diverse weather conditions or nighttime scenarios. Including these factors could enhance the robustness of the adversarial patterns. Lastly, as~detection models are retrained or updated, a~pattern optimized for one version may lose effectiveness over time. Exploring continual or online adversarial training strategies could sustain attack performance against evolving systems.
In conclusion, TACO establishes a strong foundation for adversarial camouflage on 3D vehicle models and offers a promising strategy for bypassing state-of-the-art object detectors. Addressing these areas in future work could further enhance its utility and ensure broader effectiveness in real-world applications.
}
\vspace{6pt}

\authorcontributions{ 
Conceptualization, A.D., T.V.M. and V.R.; methodology, A.D. and T.V.M.; software, A.D. and T.V.M.; resources, V.R.; data curation, A.D. and T.V.M.; writing---original draft preparation, A.D.; writing---review and editing, T.V.M. All authors have read and agreed to the published version of the manuscript.
}

\funding{This 

 research was funded by the Ministry of Culture and Innovation of Hungary from the National Research, Development and Innovation Fund, financed under the ``Nemzeti Laboratóriumok pályázati
program'' funding scheme, grant number~2022-2.1.1-NL-2022-00012.}
 
\dataavailability{The datasets presented in this study are part of an ongoing research project and are therefore not readily available. For~access requests, please contact Adonisz Dimitriu at dimitriu.adonisz@techtra.hu.}

\acknowledgments{We 
 would like to acknowledge Eszter Fülöp for her support in developing the figures and offering valuable design advice. Her expertise in visual representation was key to the overall presentation of this~work.}

\conflictsofinterest{The authors declare no conflicts of interest. The~funders had no role in the design of the study; in the collection, analyses, or~interpretation of data; in the writing of the manuscript; or in the decision to publish the~results.}

\begin{adjustwidth}{-\extralength}{0cm}

\reftitle{References}

\PublishersNote{}
\end{adjustwidth}
\end{document}